\documentclass[conference,romanappendices]{IEEEtran}
\IEEEoverridecommandlockouts
\usepackage{cite}
\usepackage{amsmath,amssymb,amsfonts}
\usepackage{algorithmic}
\usepackage{graphicx}
\usepackage{subfig}
\usepackage{textcomp}
\usepackage{balance}
\usepackage{xcolor}
\usepackage{pgfplots}
\pgfplotsset{compat=1.18}
\usepackage{ifthen}
\usepackage{url}
\def\BibTeX{{\rm B\kern-.05em{\sc i\kern-.025em b}\kern-.08em
    T\kern-.1667em\lower.7ex\hbox{E}\kern-.125emX}}

\usepackage{color}

\DeclareMathOperator*{\argmin}{arg\,min}

\definecolor{dark_blue}{RGB}{64,5,157}   
\definecolor{purple}{RGB}{143,13,164}   
\definecolor{pink}{RGB}{224,100,99} 
\definecolor{yellow}{RGB}{253,207,37}

\newcommand{\usetikz}{1}  
\newcommand{\figheight}{200px}
\newcommand{\legendopacity}{0.75}

\begin{document}

\title{An Active Inference Model of Covert and Overt Visual Attention\\
\thanks{This research has been supported by the H2020 project AIFORS under Grant Agreement No 952275 \\ \textsuperscript{1}University of Zagreb Faculty of Electrical Engineering and Computing, Croatia; Correspondence: tin.misic@fer.hr, ivan.markovic@fer.hr}
}

\author{\IEEEauthorblockN{Tin Mišić, Karlo Koledić, Fabio Bonsignorio, Ivan Petrović, and Ivan Marković\textsuperscript{1}}
}

\maketitle

\begin{abstract}
The ability to selectively attend to relevant stimuli while filtering out distractions is essential for agents that process complex, high-dimensional sensory input. This paper introduces a model of covert and overt visual attention through the framework of active inference, utilizing dynamic optimization of sensory precisions to minimize free-energy. The model determines visual sensory precisions based on both current environmental beliefs and sensory input, influencing attentional allocation in both covert and overt modalities. To test the effectiveness of the model, we analyze its behavior in the Posner cueing task and a simple target focus task using two-dimensional(2D) visual data. Reaction times are measured to investigate the interplay between exogenous and endogenous attention, as well as valid and invalid cueing. The results show that exogenous and valid cues generally lead to faster reaction times compared to endogenous and invalid cues. Furthermore, the model exhibits behavior similar to inhibition of return, where previously attended locations become suppressed after a specific cue-target onset asynchrony interval. Lastly, we investigate different aspects of overt attention and show that involuntary, reflexive saccades occur faster than intentional ones, but at the expense of adaptability.
\end{abstract}


\section{Introduction}
Attention as a cognitive process allows agents to selectively focus on specific stimuli while ignoring others. 
This ability helps humans avoid sensory overload, and as robots acquire more complex sensory channels  it could help decrease the computational load required to perform in daily tasks, such as object tracking and visual search, as well as social interactions \cite{lanillos1, lanillos2, lanillos3}. 
Attention is often separated into top-down, or goal-driven attention, and bottom-up or stimulus-driven attention, with some theories including hysteresis as a third component \cite{shomstein}. 
Top-down attention bilaterally activates dorsal posterior parietal and frontal regions of the brain, while bottom-up attention activates the right-lateralized ventral system, with the dorsal frontoparietal system combining the two into a ``salience map'' during visual search \cite{corbetta, mengotti}. 
Furthermore, visual attention is separated into overt and covert attention\cite{kulke, blair}, with overt attention involving saccadic eye movements to the attentional target, and covert attention referring to attention shifts to the target while the eyes remain fixated elsewhere. 
Multiple approaches exist to model attention, more numerous being those that are based on  Bayesian inference \cite{feldman,parr18,parr17,mirza,parr19,parvizi,spratling,itti}. 
While previous studies have modeled visual attention and active saccades in visual search \cite{parr17,parr18,mirza,feldman}, the integration of visual attention and bottom-up action from raw two-dimensional visual data within the active inference framework remains unexplored.
This is particularly important in robotics, as vision is a fundamental sensory modality, with images serving as a primary source of perceptual input for decision-making and interaction with the environment.

\begin{figure}[!t]
    \centering
    \includegraphics[width=1\linewidth]{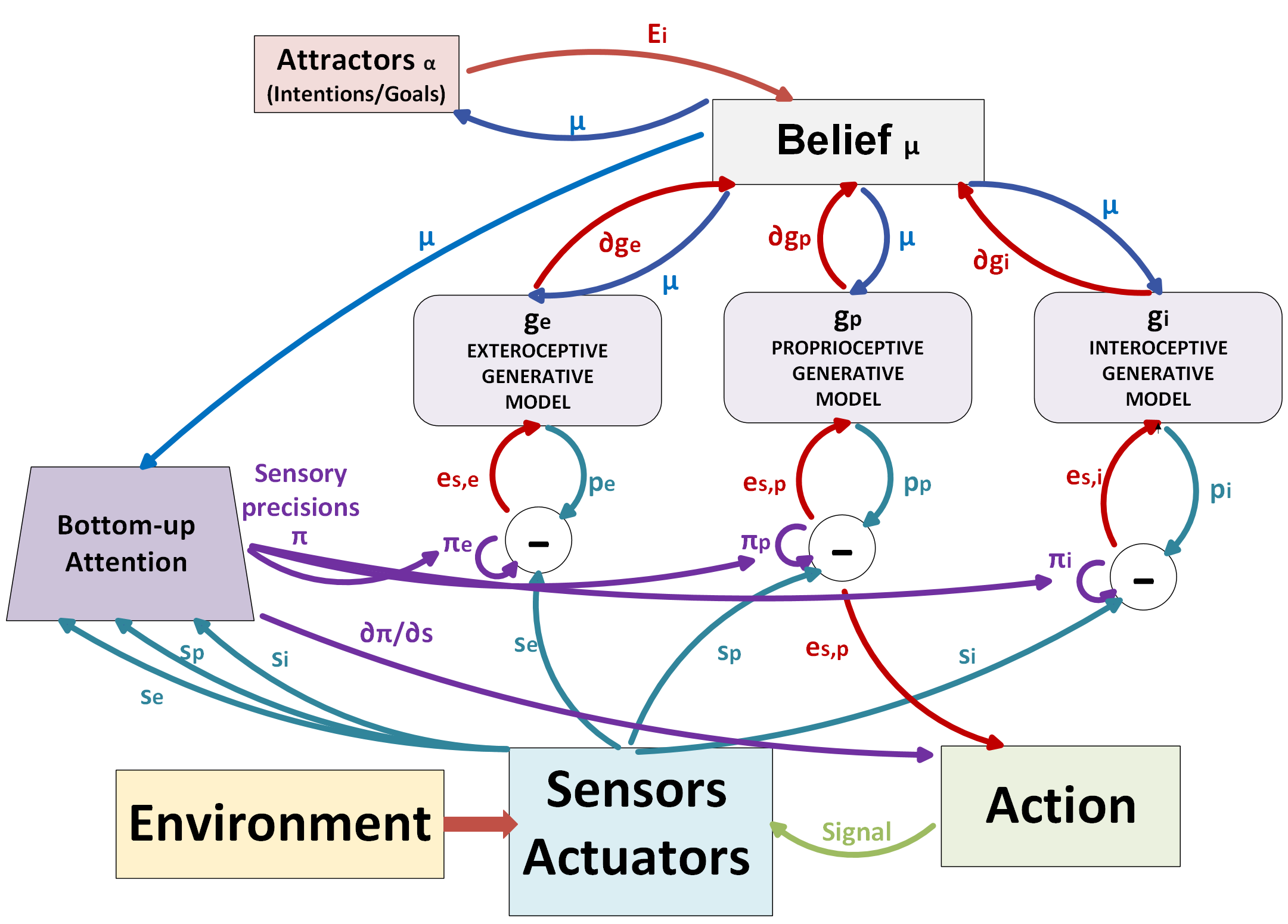}
    \caption{At the core of the proposed model are the beliefs about the causes of sensory inputs. These beliefs and action signals are updated through attractor goals and error updates to minimize free-energy. The dedicated bottom-up attention module regulates attention through dynamic sensory precisions. }
    \label{fig:model}
\end{figure}

Visual attention and its models are most often tested using the Posner cueing task, i.e., the Posner paradigm. 
The Posner cueing task is an experimental paradigm used to study covert visual attention \cite{posner1, posner2}. 
Participants are asked to fixate on a central point while a cue directs attention to a location where a target may appear. 
The cue can either be endogenous -- meaning that attention is voluntarily guided based on symbolic cues (e.g., an arrow pointing left or right), or exogenous -- meaning that attention is automatically drawn by a sudden, peripheral stimulus (e.g. a bright flash or a flickering box). 
Endogenous cueing is considered to be top-down because it requires cognitive processing and active interpretation of the cue, while exogenous cueing is considered to be bottom-up because it does not require conscious interpretation. 
Reaction times and accuracy are measured to assess how cues influence attentional shifts.

Through the original Posner paradigm \cite{posner1, posner2} and its variations, valuable insights have been gained about attentional processes. 
Covert attentional shifts to a target area occur prior to any eye movement \cite{posner2, peterson}, and valid cues produce faster responses than invalid cues\cite{posner1, posner2}. 
Exogenous cues were shown to produce faster reaction times than endogenous cues \cite{jonides, cheal}, showing that bottom-up attention is faster because it requires no conscious processing. 
The question of weather attentional selection is object-based or location-based has also been thoroughly researched, and the consensus is that both types are not mutually exclusive, but are dependent on the current task \cite{vecera, egly, reppa}. 
Research supporting location-based attention has shown that the distance from the focus point plays a role in reaction time, with reaction times increasing as target eccentricity increased \cite{klein, downing, carrasco}.

In this paper we propose a model of visual attention, shown in Fig.~\ref{fig:model}, viewed through the lens of active inference\cite{parr_aif} -- a computational approach derived from the free-energy principle (FEP). 
According to the FEP, systems adapt and act in a way that minimizes their free-energy \cite{Friston2006}. 
Free-energy is a concept borrowed from physics, statistics, and information theory that limits the surprise on a sample of data given a generative model. 
This principle helps to explain how biological systems resist the natural tendency to disorder, and their action, perception, and learning processes \cite{Friston2010}. 
In the FEP, attention is theoretically achieved by optimizing sensory precisions, their parameters, and mutual precision weighing \cite{feldman, parr18, parr17, parr19, mirza, kanai, parvizi}.
Biased competition and endogenous/exogenous attention have been studied in this context, and the precision optimization produces behaviors similar to human attention \cite{feldman, spratling}.

The contribution of this paper is an active inference model of overt and covert visual attention by investigating precision optimization for visual data and how it generates endogenous/exogenous attention and action control.
The proposed model includes both top-down and bottom-up visual attention, as well as covert and overt shifts in attention. 
These properties are demonstrated through the Posner cueing task and a simple target focus task on visual 2D data. 
A variational auto-encoder (VAE) was used for the visual generative model, and model training and experiments were done in the Gazebo simulator in the Robot Operating System (ROS).

The paper is organized as follows. 
In Sec.~\ref{methods} we give an overview of the theoretical background and elaborate the proposed approach that is based on free-energy minimization with 2D precision optimization and overt saccades through active inference. 
Section~\ref{results} shows the results of the Posner cueing tasks and active attention trials.
Section~\ref{discussion} provides the discussion of the results while Sec.~\ref{conclusion} concludes the paper and provides directions for future work.

\section{Proposed Method} \label{methods}

\subsection{Free-energy Minimization}
Free-energy is defined as the negative evidence lower bound (ELBO), or as the sum of the Kullback-Leibler (KL) divergence and the surprise\cite{Friston2006,Friston2010,feldman}:
\begin{equation}
    \begin{aligned}
        F(\boldsymbol{z},\boldsymbol{s}) & = - \mathcal{L}(q) \\
               & = D_{KL}[q(\boldsymbol{z})||p(\boldsymbol{z}|\boldsymbol{s})] - \ln p(\boldsymbol{s}),
    \end{aligned}
    \label{eq:free-energy}
\end{equation}
where $\boldsymbol{z}$ and $\boldsymbol{s}$ represent latent system  states and sensory observations, respectively, while the KL-divergence is computed between the posterior \(p(\boldsymbol{z}|\boldsymbol{s})\) and the approximate variational density \(q(\boldsymbol{z})\). 
Given that, the surprise is defined as the negative log-probability of an outcome $-\ln p(\boldsymbol{s})$. 
If the variational density \(q(\boldsymbol{z})\) is assumed to factor into Gaussian probability density functions (pdfs) \cite{Friston2006,priorelli,feldman}: 
\begin{equation}
    q(\boldsymbol{z}) = \prod_i q(\boldsymbol{z}_i) = \prod_i \mathcal{N}(\boldsymbol{\mu}_i, \boldsymbol{\Pi}_i^{-1}),
\end{equation}
the free-energy then becomes dependent only on the most probable hypotheses, {beliefs} \(\boldsymbol{\mu_i}\), and precision matrices \(\boldsymbol{\Pi}_i\) of the latent system states \(\boldsymbol{z}\)\cite{priorelli,feldman}:

\begin{equation}
    \begin{aligned}
        F(\boldsymbol{\mu},\boldsymbol{s}) &= - \ln p(\boldsymbol{s},\boldsymbol{\mu}) + C\\
        &= - \ln p(\boldsymbol{s} | \boldsymbol{\mu}) - \ln p(\boldsymbol{\mu}) + C.
    \end{aligned}
    \label{eq:femu}
\end{equation}

Furthermore, sensory observations \(\boldsymbol{s}\) and beliefs \(\boldsymbol{\mu}\) are defined in the context of hierarchical dynamic models\cite{priorelli,Friston2006,Friston2010,feldman}:
\begin{equation}
    \begin{aligned}
        & \boldsymbol{\tilde s} = \boldsymbol{\tilde g}(\boldsymbol{\tilde \mu}) + \boldsymbol{w}_s \\
        & D\boldsymbol{\tilde \mu} = \boldsymbol{\tilde f}(\boldsymbol{\tilde \mu}) + \boldsymbol{w}_\mu.
    \end{aligned}
\end{equation}
Here, \(\boldsymbol{\tilde \mu}\) indicates generalized coordinates of beliefs with multiple temporal orders, \(\boldsymbol{\tilde \mu} = \{\boldsymbol{\mu}, \boldsymbol{\mu}^{\prime},\boldsymbol{\mu}^{\prime\prime},\cdots\}\), which allow for a richer approximation of the environment dynamics, \(D\) stands for the differential shift operator \(D\boldsymbol{\tilde \mu} = \{\boldsymbol{\mu}^{\prime},\boldsymbol{\mu}^{\prime\prime},\cdots\}\) in the generalized equation of system dynamics \(\boldsymbol{\tilde f}(\boldsymbol{\tilde \mu})\), while $\boldsymbol{\tilde g}(\boldsymbol{\tilde \mu})$ is the sensor model that maps current beliefs to sensory observations.
The amplitudes of random fluctuations \(\boldsymbol{w}_s\) and \(\boldsymbol{w}_\mu\) are state dependent and are defined as Gaussian pdfs with covariances \(\boldsymbol{\Sigma_s}\) and \(\boldsymbol{\Sigma_\mu}\), respectively\cite{priorelli,feldman}: 
\begin{equation}
    \begin{aligned}
        & \boldsymbol{w_s} \sim \mathcal{N}(\boldsymbol{\mu}_i, \boldsymbol{\Sigma}_s(\boldsymbol{z},\boldsymbol{s},\boldsymbol{\gamma}))\\
        & \boldsymbol{w_\mu} \sim \mathcal{N}(\boldsymbol{\mu}_i, \boldsymbol{\Sigma}_\mu(\boldsymbol{z},\boldsymbol{s},\boldsymbol{\gamma})).
    \end{aligned}
\end{equation}
The precision matrices \(\boldsymbol{\Pi}_i\) are the inverses of these covariances, \(\boldsymbol{\Pi}_i := \boldsymbol{\Pi}_i(\boldsymbol{z},\boldsymbol{s},\boldsymbol{\gamma}) = \boldsymbol{\Sigma}_i(\boldsymbol{z},\boldsymbol{s},\boldsymbol{\gamma})^{-1}\), with precision parameters \(\boldsymbol{\gamma}\) that control the amplitudes\cite{feldman,spratling}. 
The precisions are dynamic and depend on the current states and sensory input. 
It is through optimization of precisions and their parameters that attention is achieved\cite{feldman, parr18, parr17, parr19, mirza, kanai, parvizi}.

\subsection{Perceptual and Active Inference}

Perception, action, and learning can all be optimized through the minimization of free-energy.
In this paper we only consider perception and action, and leave the learning processes of attention for future work. 
Action and beliefs are optimized through gradient descent\cite{parr_aif,Friston2006,Friston2010,priorelli}:
\begin{equation}
\begin{aligned}
    & \boldsymbol{\dot {\tilde \mu}} - D\boldsymbol{\tilde \mu} = - \partial_{\tilde \mu} F(\boldsymbol{\tilde \mu},\boldsymbol{\tilde s})\\
    & \boldsymbol{\dot a} =  - \partial_{\boldsymbol{a}} F(\boldsymbol{\tilde \mu},\boldsymbol{\tilde s}).
\end{aligned}
\label{eq:mu_a_optim}
\end{equation}
The likelihood and prior in (\ref{eq:femu}) also become generalized and can be partitioned within and across temporal orders \(d\), respectively\cite{priorelli}:
\begin{equation}
    \begin{aligned}
        p(\boldsymbol{\tilde s} | \boldsymbol{\tilde \mu}) &=\underset{d} {\prod} p(\boldsymbol{s}^{[d]} | \boldsymbol{\mu}^{[d]})\\
        p(\boldsymbol{\tilde \mu}) &=\underset{d} {\prod} p(\boldsymbol{\mu}^{[d+1]} | \boldsymbol{\mu}^{[d]}). 
    \end{aligned}
\end{equation}
These partitions are also assumed to take the following Gaussian pdf form:
\begin{equation}
    \begin{aligned}
        p(\boldsymbol{s}^{[d]} | \boldsymbol{\mu}^{[d]}) & =  \frac{|\boldsymbol{\Pi_s}^{[d]}|^{\frac{1}{2}}}{\sqrt{\left(2\boldsymbol{\pi}\right)^L} } \exp\left({-\frac{1}{2} {\boldsymbol{e}_s^{[d]}}^T \boldsymbol{\Pi_s}^{[d]}\boldsymbol{e}_s^{[d]} } \right) \\
        p(\boldsymbol{\mu}^{[d+1]} | \boldsymbol{\mu}^{[d]}) & =  \frac{|\boldsymbol{\Pi_{\mu}}^{[d]}|^{\frac{1}{2}}}{\sqrt{\left(2\boldsymbol{\pi}\right)^M} } \exp\left({-\frac{1}{2} {\boldsymbol{e}_{\mu}^{[d]}}^T \boldsymbol{\Pi_\mu}^{[d]}\boldsymbol{e}_\mu^{[d]} } \right),
    \end{aligned}
\end{equation}
where \(L\) and \(M\) are the respective dimensions of sensory observations $\boldsymbol{s}$ and internal beliefs $\boldsymbol{\mu}$.
Therein, \(\boldsymbol{e}_s^{[d]}\) and \(\boldsymbol{e}_\mu^{[d]}\) represents sensory and system dynamics prediction errors:
\begin{equation}
    \begin{aligned}
        &\boldsymbol{e}_s^{[d]} = \boldsymbol{s}^{[d]} - \boldsymbol{g}^{[d]}(\boldsymbol{\mu}^{[d]}) = \boldsymbol{s}^{[d]} -\boldsymbol{p}^{[d]}\\
        &\boldsymbol{e}_\mu^{[d]} = \boldsymbol{\mu}^{[d+1]} - \boldsymbol{f}^{[d]}(\boldsymbol{\mu}^{[d]}),
    \end{aligned}    
    \label{eq:sensory_err}
\end{equation}
where \(\boldsymbol{p}^{[d]} = \boldsymbol{g}^{[d]}(\boldsymbol{\mu}^{[d]})\) are sensory predictions generated by the generative sensor model. Note that in our case the system dynamics model is defined through flexible intentions \(\boldsymbol{h}^{(k)}\)\cite{priorelli}, where for each intention $k\in(0,K-1)$:
\begin{equation}
    \boldsymbol{f}^{(k)}(\boldsymbol{\mu}) = l\cdot\boldsymbol{E}_i^{(k)} + \boldsymbol{w}^{(k)}_\mu = l\cdot(\boldsymbol{h}^{(k)} - \boldsymbol{\mu}) + \boldsymbol{w}^{(k)}_\mu,
    \label{eq:system_dynamics}
\end{equation}
with $l$ being the gain of intention errors $\boldsymbol{E}_i^{(k)}$. The implementation of the generative sensor models $\boldsymbol{g}^{[d]}$ is presented in subsection \ref{implementation}.

\subsubsection{Belief update}

With state- and sensory-dependent precisions, the belief update takes the following form: 
\begin{equation}
    \begin{aligned}
        \boldsymbol{\dot {\tilde \mu}} = &  D\boldsymbol{\tilde \mu} + \frac{\partial \boldsymbol{\tilde g}}{\partial \boldsymbol{\tilde \mu}}^T {\boldsymbol{\tilde \Pi}_s} \boldsymbol{\tilde e}_s + \frac{\partial \boldsymbol{\tilde f}}{\partial \boldsymbol{\tilde \mu}}^T {\boldsymbol{\tilde \Pi}_\mu} \boldsymbol{\tilde e}_\mu - D^T {\boldsymbol{\tilde \Pi}_\mu} \boldsymbol{\tilde e}_\mu\\
        & + \frac{1}{2} \mathrm{Tr}\left[\boldsymbol{\tilde{\Pi}_s^{-1}}\frac{\partial\boldsymbol{\tilde{\Pi}_s}}{\partial \boldsymbol{\tilde \mu}}\right] - \frac{1}{2} {\boldsymbol{\tilde e}_{s}^T \frac{\partial\boldsymbol{\tilde{\Pi}_s}}{\partial \boldsymbol{\tilde \mu}}\boldsymbol{\tilde e}_s} \\
        & + \frac{1}{2} \mathrm{Tr}\left[\boldsymbol{\tilde{\Pi}_\mu^{-1}}\frac{\partial\boldsymbol{\tilde{\Pi}_\mu}}{\partial \boldsymbol{\tilde \mu}}\right] - \frac{1}{2} {\boldsymbol{\tilde e}_{\mu}^T} \frac{\partial\boldsymbol{\tilde{\Pi}_\mu}}{\partial \boldsymbol{\tilde \mu}}\boldsymbol{\tilde e}_\mu,
    \end{aligned}
    \label{eq:belief_update}
\end{equation}
with \(\mathrm{Tr}\) being the trace of a matrix.
The terms that comprise the belief update equation are:
\begin{itemize}
    \item \(\frac{\partial \boldsymbol{\tilde g}}{\partial \boldsymbol{\tilde \mu}}^T {\boldsymbol{\tilde \Pi}_s} \boldsymbol{\tilde e}_s\) : likelihood error computed at the sensory level, representing the free-energy gradient of the likelihood relative to the belief \(\boldsymbol{\tilde \mu}^{[d]}\) in (\ref{eq:sensory_err})
    \item \(\frac{\partial \boldsymbol{\tilde f}}{\partial \boldsymbol{\tilde \mu}}^T {\boldsymbol{\tilde \Pi}_\mu} \boldsymbol{\tilde e}_\mu\) : backward error from the next temporal order, representing the free-energy gradient relative to the belief \(\boldsymbol{\tilde \mu}^{[d+1]}\) in (\ref{eq:sensory_err})
    \item \(- D^T {\boldsymbol{\tilde \Pi}_\mu} \boldsymbol{\tilde e}_\mu\) : forward error coming from the previous temporal order, representing the free-energy gradient relative to the belief \(\boldsymbol{\tilde \mu}^{[d]}\) in (\ref{eq:sensory_err})
    \item \(\frac{1}{2} \mathrm{Tr}\left[\boldsymbol{\tilde{\Pi}_s^{-1}}\frac{\partial\boldsymbol{\tilde{\Pi}_s}}{\partial \boldsymbol{\tilde \mu}}\right] - \frac{1}{2} {\boldsymbol{\tilde e}_{s}^T \frac{\partial\boldsymbol{\tilde{\Pi}_s}}{\partial \boldsymbol{\tilde \mu}}\boldsymbol{\tilde e}_s}\): free-energy gradients from the sensory precisions, serves as bottom-up attention
    \item \(\frac{1}{2} \mathrm{Tr}\left[\boldsymbol{\tilde{\Pi}_\mu^{-1}}\frac{\partial\boldsymbol{\tilde{\Pi}_\mu}}{\partial \boldsymbol{\tilde \mu}}\right] - \frac{1}{2} {\boldsymbol{\tilde e}_{\mu}^T} \frac{\partial\boldsymbol{\tilde{\Pi}_\mu}}{\partial \boldsymbol{\tilde \mu}}\boldsymbol{\tilde e}_\mu\): free-energy gradients from the system dynamics precisions, serves as top-down attention.
\end{itemize}

\subsubsection{Action update}

Action is also updated through the minimization of free-energy\cite{Friston2006,Friston2010,priorelli,parr_aif}:
\begin{equation}
    a = \argmin_a F(\boldsymbol{\mu},\boldsymbol{s}),
\end{equation}
with the action update taking the following form:
\begin{equation}
    \begin{aligned}
        \dot{a} = &-\partial_aF(\boldsymbol{\mu},\boldsymbol{s}) = - \frac{\partial \boldsymbol{\tilde{s}}}{\partial a}^T{\boldsymbol{\tilde \Pi}_s} \boldsymbol{\tilde e}_s \\
        &+  \frac{1}{2} \mathrm{Tr}\left[\boldsymbol{\tilde{\Pi}_s^{-1}}\frac{\partial\boldsymbol{\tilde{\Pi}_s}}{\partial \boldsymbol{\tilde s}}\right]\frac{\partial \boldsymbol{\tilde{s}}}{\partial a} - \frac{1}{2} {\boldsymbol{\tilde e}_{s}^T\frac{\partial\boldsymbol{\tilde{\Pi}_s}}{\partial \boldsymbol{\tilde s}}\boldsymbol{\tilde e}_s}\frac{\partial \boldsymbol{\tilde{s}}}{\partial a},
    \end{aligned}
    \label{eq:action_update}
\end{equation}
with bottom-up attention components in relation to sensory input, analogous to those in relation to belief in (\ref{eq:belief_update}). 
These control signals act as reflexive saccades \cite{walker, kauffmann}.
The gradient $\frac{\partial \boldsymbol{\tilde{s}}}{\partial a}$ is an inverse mapping from sensory data to actions, which is usually considered a "hard problem"\cite{Friston2010_2}.

The implementations of all gradients in terms of belief, action and sensory input are elaborated in Appendix \ref{appendix}.

\section{Results} \label{results}

\subsection{Implementation of the proposed model}\label{implementation}

The graphical representation of the developed model\footnote{The implemented model is available at: \url{https://github.com/TinMisic/AIF---visual-attention/tree/ICDL}} can be seen in Fig. \ref{fig:model}. The current belief \(\boldsymbol{\mu}\) is passed as input to exteroceptive, proprioceptive, and interoceptive generative models.
The predictions \(\boldsymbol{p}\) of these models are compared to the actual sensory input $\boldsymbol{s}$ and the prediction errors \(\boldsymbol{e_s}\) are used to drive action, as well as to update the current beliefs. 
The generative models for proprioceptive (camera pitch and yaw) and interoceptive (symbolic cue signals) sensory input are trivial identity matrices, while the generative model for the exteroceptive visual sensory input is the decoder of a disentangled variational auto-encoder (VAE). 
The VAE has been trained to disentangle the position of the target in the image, as well as the target's presence in the image. 
This disentanglement simplifies the conversion from intrinsic image coordinates to extrinsic camera orientation angles.
The VAE architecture, training and latent space encoding are elaborated in Appendix \ref{vae}.

The belief state is composed of the following components: 
\begin{itemize}
    \item \textbf{Symbolic cue belief} -- interoceptive endogenous cues will present the cue position on the image, and this belief should mirror that from the sensory input
    \item \textbf{Camera orientation belief} -- proprioceptive belief over the extrinsic pitch and yaw angles of the camera viewing the environment
    \item \textbf{Visual belief} - an encoding of the exteroceptive visual input, disentangled to encode the target position and presence so they are easily interpreted
    \item \textbf{Covert attention belief} -  belief over the amplitude and center of a radial basis function (RBF) used to calculate the visual sensory precisions.
\end{itemize}
The sensory data and belief shapes are elaborated in Appendix \ref{sensory_belief}.
The beliefs are updated through bottom-up prediction error gradients, as well as through top-down attractors \(\boldsymbol{\alpha}\) generated from the current beliefs, according to the flexible intentions theory proposed in \cite{priorelli}. 
These goal-directed intentions encourage action through the proprioceptive camera orientation, as well as covert attention through the shifts of the RBF center and amplitude.

\begin{figure}[!t]
    \centering
    \subfloat[Visual sensory precision matrix]{%
        \includegraphics[height=0.35\linewidth]{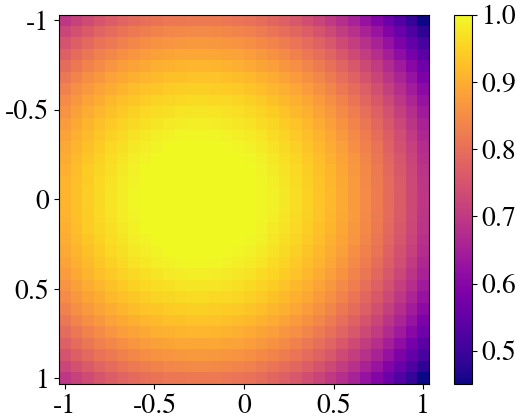}%
        \label{fig:precision}
    }
    \hfill
    \subfloat[Sensory precision free-energy gradient, $\boldsymbol{\dot\mu_u} = -0.839$]{%
        \includegraphics[height=0.35\linewidth]{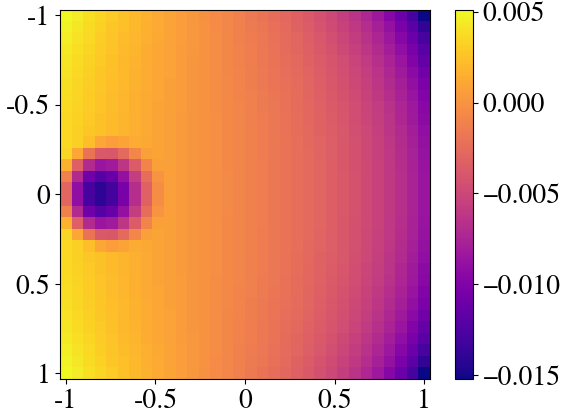}%
        \label{fig:update}
    }
    \caption{The center of the RBF is (-0.25, 0.0), while the error appears at (-0.75, 0.0). The $u$-component of the RBF center is pushed toward the error with the update $\boldsymbol{\dot\mu_u} = -0.839$.}
    \label{fig:rbf}
\end{figure}

Sensory precision \(\boldsymbol{\Pi_s}\) for the visual input is dynamic and calculated based on the current overt attention belief and sensory input. We assume that there is no correlation between individual pixels, so \(\boldsymbol{\Pi_s}\) is defined as:
\begin{equation}
\boldsymbol{\Pi_s} = \begin{bmatrix}
\pi_1(\boldsymbol{\mu},\boldsymbol{s}) & 0  & \cdots & 0 \\
0 & \pi_2(\boldsymbol{\mu},\boldsymbol{s})  & \cdots & 0 \\
\vdots & \vdots  & \ddots & \vdots \\
0 & 0 & \cdots & \pi_L(\boldsymbol{\mu},\boldsymbol{s})
\end{bmatrix}_{L \times L},
\end{equation}
where \(L = 32\times 32\left(\times 3\right)\) is the dimensionality of the visual data. We further assume that the individual precision functions \(\pi_i(\boldsymbol{\mu},\boldsymbol{s})\) are determined by RBFs based on the covert attention center and the presence of a target-specific property, in our case the color red:

\begin{equation}
    \begin{aligned}
        &\pi_i(\boldsymbol{\mu},\boldsymbol{s}) = \pi(x,y,\boldsymbol{\mu},\boldsymbol{s}) =\\
        &\frac{\mu_{amp}}{2}\left(\ln\left(-\frac{(x - \mu_u)^2 + (y - \mu_v)^2}{b^2} + 1\right)+ c\right)\\
        &+\frac{1}{2}\left(\ln\left(-\frac{(x - \boldsymbol{r_u}(\boldsymbol{s}))^2 + (y - \boldsymbol{r_v}(\boldsymbol{s}))^2}{b^2} + 1\right)+ c\right),
    \end{aligned}
    \label{eq:little_pi}
\end{equation}
where \(\left[\mu_{amp},\mu_{u},\mu_{v}\right]\) are covert attention beliefs, \(\left[\boldsymbol{r_u}(\boldsymbol{s}),\boldsymbol{r_v}(\boldsymbol{s})\right]\) is the centroid of the biggest red object.
The parameters of the precision function, \( b = 2.6 \) and \( c = 1 \), are empirically chosen to ensure that the RBF values span from 0 to 1 across the image area.
The shape of the RBF was chosen so that the belief update pushes the covert attention toward the area of the image with the highest error, while a Gaussian RBF would push it away from the error. The sensory precision matrix generated by this RBF and the resulting free-energy gradient caused by a prediction error can be seen in Fig. \ref{fig:rbf}. 
The precision nevertheless decreases the further a point is from the focus center, mimicking human foveation \cite{downing, carrasco}.

\subsection{Simulating the Posner Cueing Task}

\begin{figure}[!t]
    \centering
    \includegraphics[width=0.7\linewidth]{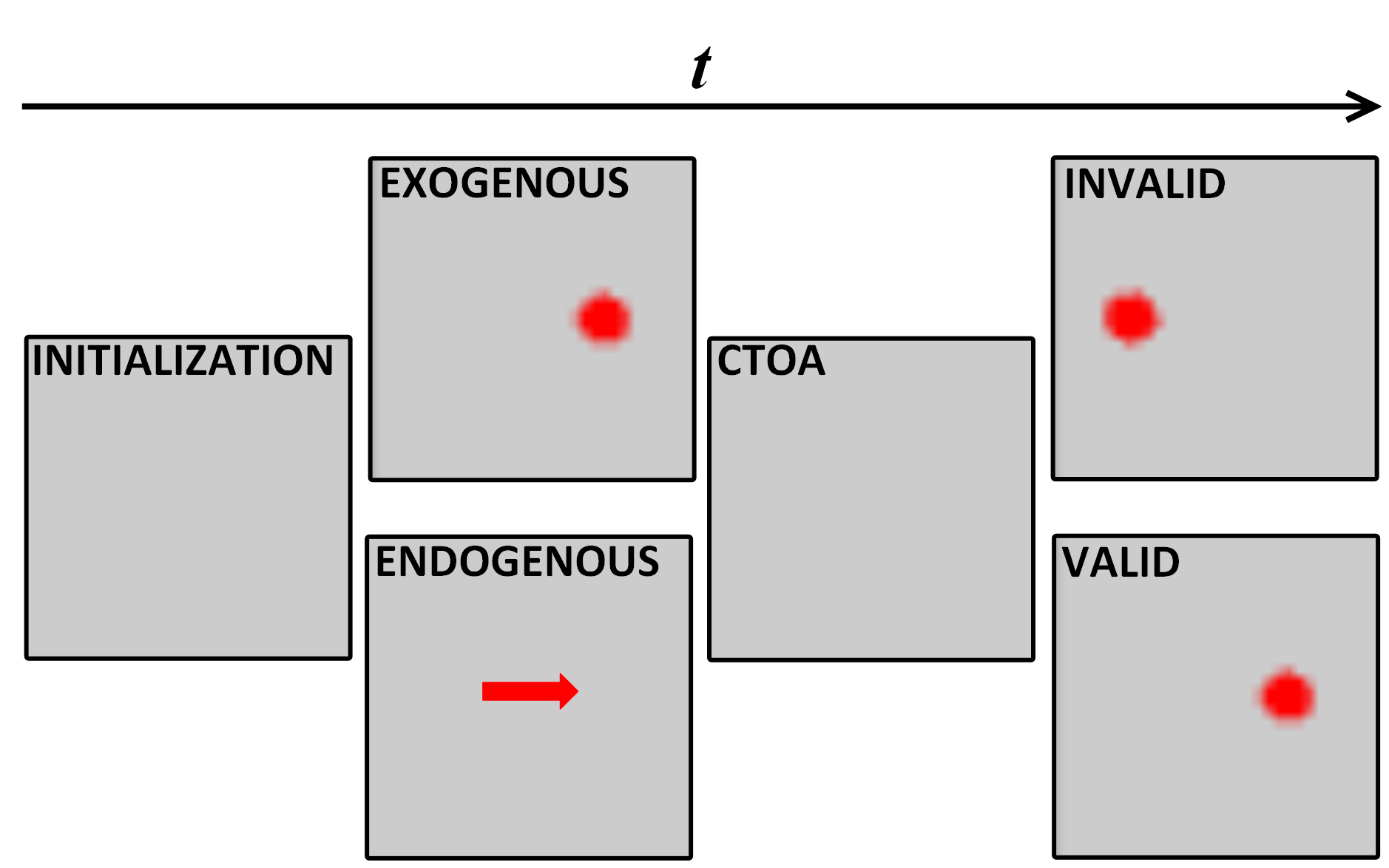}
    \caption{Trial sequence of events. The model is first initialized for 10 steps, then a cue appears for 50 simulation steps. The cue is then removed for a variable interval, known as cue-target onset asynchrony (CTOA). After that the target appears until it is detected by the model or 1000 steps have passed.}
    \label{fig:setup}
\end{figure}

The Posner cueing task is used to demonstrate the proposed model's exogenous and endogenous covert attention.
The model's sensory inputs are its current camera orientation, a symbolic cue signal and visual data of an empty scene in which a red sphere might appear as a target. 
Note that the endogenous cue is given through the interoceptive sensory channel, not as an arrow in the visual channel as illustrated in Fig.\ref{fig:setup}.
We performed four variations of the cueing task, for both endogenous and exogenous cueing in valid and invalid settings. 
The endogenous cue is given through the symbolic cue signal which has to be processed into an intention that moves both the center of covert attention and the belief over the sphere's position. 
The exogenous cue is a brief appearance of the target object, which moves the center of covert attention through the bottom-up free-energy gradient from the sensory precision, and the belief over the sphere's position through the likelihood error from the VAE. 
A valid cue setting is when the target appears at the same position as the cue, and an invalid cue setting when the target appears at a position opposite of the cue with respect to the central focus point. 

For each of the four task variations, $N=200$ trials were conducted.
For each trial, the position of the target is randomly generated with varying distance from the focus point. 
Fig. \ref{fig:setup} shows the sequence of events in a single trial. 
As this cueing task is meant to test covert attention, overt attention through action signals was disabled. 

\begin{figure}[!t]
    \centering
    \ifthenelse{\usetikz=1}{
        \begin{tikzpicture}
            \begin{axis}[
                height=\figheight,
                width=\linewidth,
                xlabel={Target distance from focus point (px)},
                ylabel={Reaction Time (steps)},
                grid=none,
                enlargelimits=0.05,
                legend style={
                    fill=white,  
                    fill opacity=\legendopacity,
                    text opacity=1,
                    font=\small,  
                    cells={anchor=west},  
                    row sep=-3pt,  
                    column sep=5pt  
                },
                legend pos=north west
                ]
                \addplot[dark_blue, mark=none, line width=2pt] table [col sep=comma, x=endovalidx, y=endovalidy] {data/posner_dist.csv};
                \addlegendentry{Endogenous-Valid};
                \addplot[dark_blue, line width=2.5pt, forget plot] coordinates {(0.1,123.22) (10.5,123.22)};
                
                \addplot[purple, dashed, mark=none, line width=2pt] table [col sep=comma, x=endoinvalidx, y=endoinvalidy] {data/posner_dist.csv};
                \addlegendentry{Endogenous-Invalid};
                \addplot[purple, dashed, line width=2.5pt, forget plot] coordinates {(0.1,142.27) (10.5,142.27)};
    
                \addplot[pink,dash dot, mark=none, line width=2pt] table [col sep=comma, x=exovalidx, y=exovalidy] {data/posner_dist.csv};
                \addlegendentry{Exogenous-Valid};
                \addplot[pink, dash dot, line width=2.5pt, forget plot] coordinates {(0.1,93.31) (10.5,93.31)};
                
                \addplot[yellow,dotted, mark=none, line width=2pt] table [col sep=comma, x=exoinvalidx, y=exoinvalidy] {data/posner_dist.csv};
                \addlegendentry{Exogenous-Invalid}
                \addplot[yellow, dotted, line width=2.5pt, forget plot] coordinates {(0.1,125.33) (10.5,125.33)};
            \end{axis}
        \end{tikzpicture}
    }{
        \includegraphics[width=1.0\linewidth]{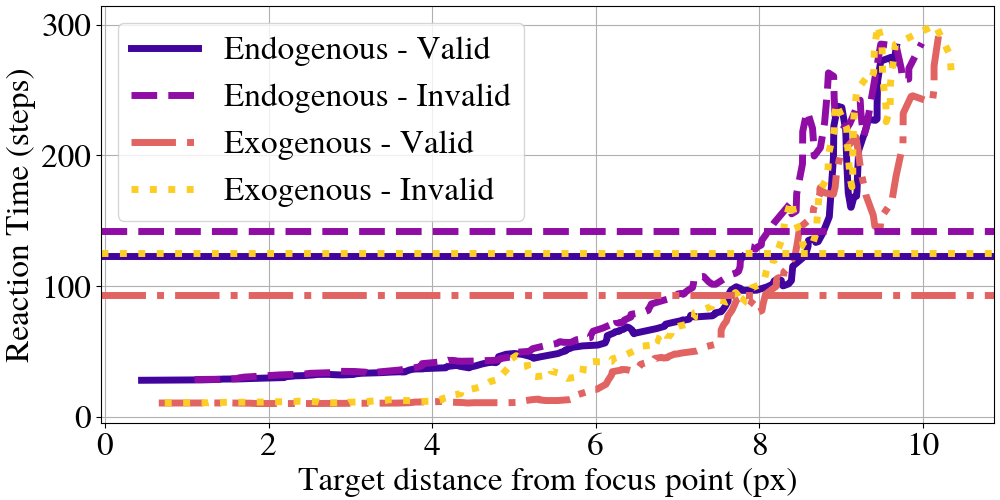}
    }
    \caption{Reaction times and their averages as a function of target distance from focus point (CTOA = 100 for each trial)}
    \label{fig:posner_dist}
\end{figure}
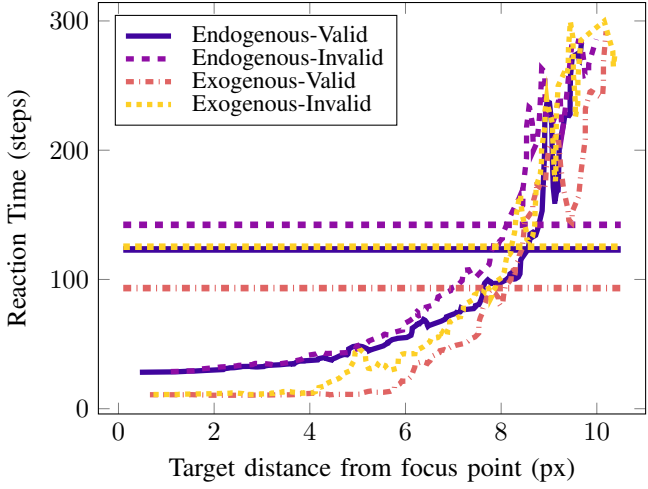

\begin{figure}[!t]
    \centering
    \ifthenelse{\usetikz=1}{
        \begin{tikzpicture}
            \begin{axis}[
                height=\figheight,
                width=\linewidth,
                xlabel={Simulation Time (steps)},
                ylabel={Distance from center (px)},
                grid=none,
                enlargelimits=0.05,
                legend style={
                    fill=white,  
                    fill opacity=\legendopacity,
                    text opacity=1,
                    font=\small,  
                    cells={anchor=west},  
                    row sep=-3pt,  
                    column sep=5pt  
                },
                legend pos=south east
                ]
    
                \addplot[red, dotted, line width=1.5pt, forget plot] coordinates {(0.1,7.15) (250,7.15)};
                \addplot[gray, dotted, line width=1.5pt, forget plot] coordinates {(11,0) (11,7.15)};
                \addplot[gray, dotted, line width=1.5pt, forget plot] coordinates {(61,0) (61,7.15)};
                \addplot[gray, dotted, line width=1.5pt, forget plot] coordinates {(161,0) (161,7.15)};
            
                \addplot[dark_blue, mark=none, line width=2pt] table [col sep=comma, x=t, y=endot] {data/posner_target.csv};
                \addlegendentry{Endogenous};
                
                \addplot[dark_blue, dashed, mark=none, line width=2pt, forget plot] table [col sep=comma, x=t, y=endof] {data/posner_target.csv};
                
                \addplot[yellow, mark=none, line width=2pt] table [col sep=comma, x=t, y=exot] {data/posner_target.csv};
                \addlegendentry{Exogenous};
                
                \addplot[yellow, dashed, mark=none, line width=2pt, forget plot] table [col sep=comma, x=t, y=exof] {data/posner_target.csv};
                
            \end{axis}
        \end{tikzpicture}
    }{
        \includegraphics[width=1.0\linewidth]{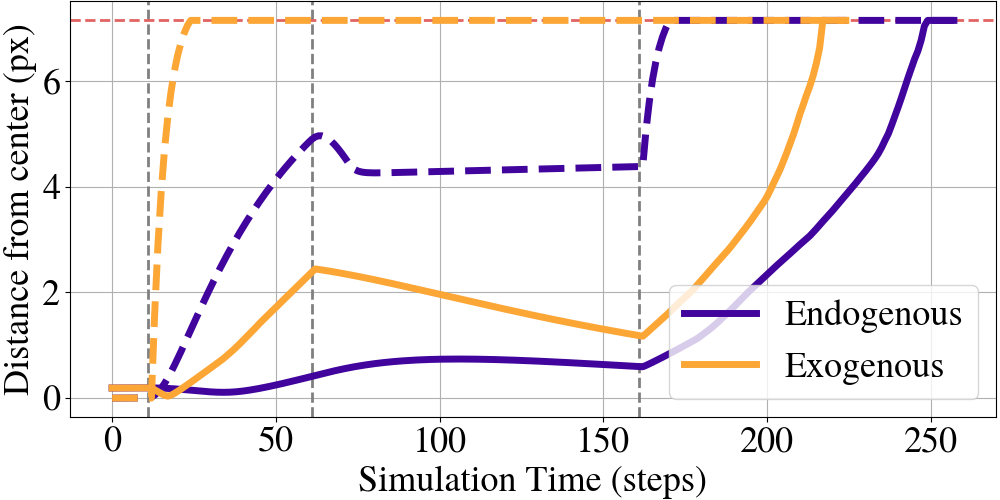}
    }
    \caption{Covert attention center (dashed lines) and sphere position beliefs (solid lines) during valid trials, for both endogenous and exogenous cues. The horizontal line is the true target distance from center, and the vertical lines indicate trial events as in Fig. \ref{fig:setup}: the cue appears at step 10, disappears at step 60, target appears at step 160.}
    \label{fig:posner_shift}
\end{figure}
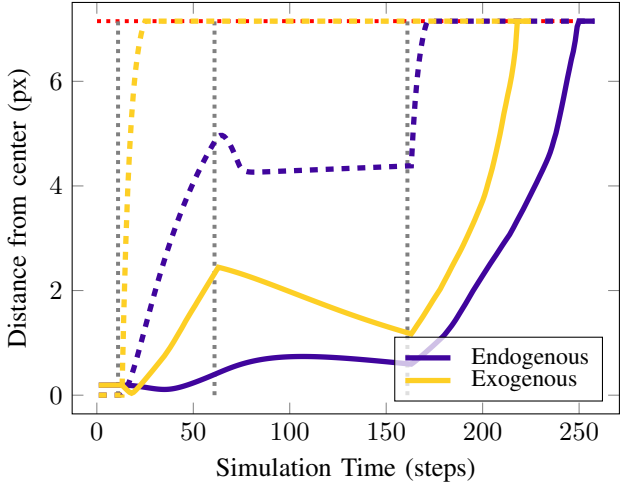

The reaction time in simulation steps as a function of distance from the focus point is shown in Fig. \ref{fig:posner_dist}, for each of the four task variations. Since the internal beliefs about the covert attention and the sphere position are easily interpretable, we can easily see the shifts of covert attention and sphere position belief for the valid task variations in Fig. \ref{fig:posner_shift}.

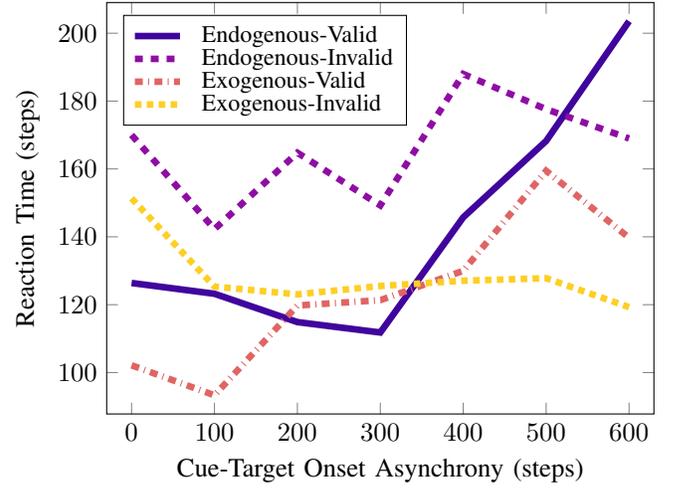
\begin{figure}[!t]
    \centering
    \ifthenelse{\usetikz=1}{
        \begin{tikzpicture}
            \begin{axis}[
                height=\figheight,
                width=\linewidth,
                enlargelimits=0.05,
                xlabel={Cue-Target Onset Asynchrony (steps)},
                ylabel={Reaction Time (steps)},
                grid=none,
                legend style={
                    fill=white,  
                    fill opacity=\legendopacity,
                    text opacity=1,
                    font=\small,  
                    cells={anchor=west},  
                    row sep=-3pt,  
                    column sep=5pt  
                },
                legend pos=north west
            ]
                \addplot[dark_blue, mark=none, line width=2.5pt] table [col sep=comma, x=ctoa, y=endovalid] {data/posner_ctoa.csv};
                \addlegendentry{Endogenous-Valid};
                
                \addplot[purple, dashed, mark=none, line width=2.5pt] table [col sep=comma, x=ctoa, y=endoinvalid] {data/posner_ctoa.csv};
                \addlegendentry{Endogenous-Invalid};
    
                \addplot[pink,dash dot, mark=none, line width=2.5pt] table [col sep=comma, x=ctoa, y=exovalid] {data/posner_ctoa.csv};
                \addlegendentry{Exogenous-Valid};
                
                \addplot[yellow,dotted, mark=none, line width=2.5pt] table [col sep=comma, x=ctoa, y=exoinvalid] {data/posner_ctoa.csv};
                \addlegendentry{Exogenous-Invalid}
            \end{axis}
        \end{tikzpicture}
    }{
        \includegraphics[width=1.0\linewidth]{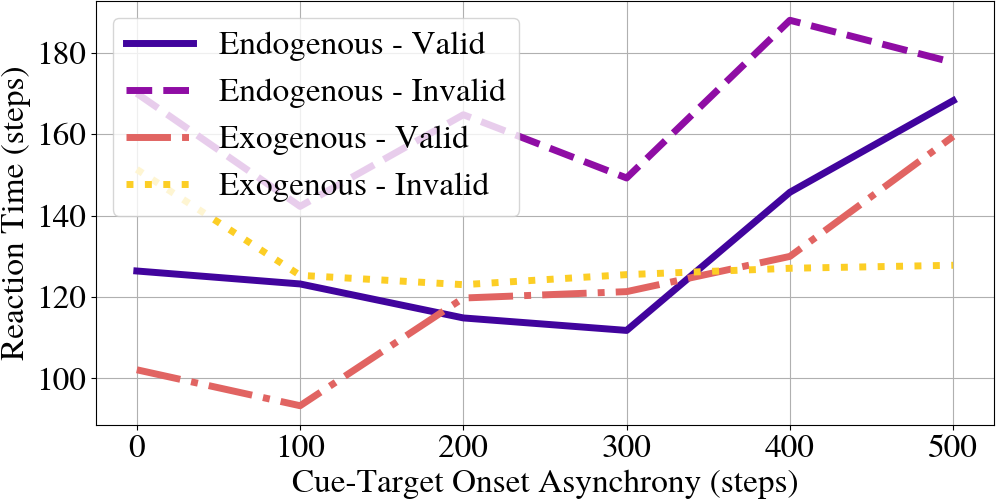}
    }
    \caption{Average trial reaction time as a function of CTOA. Results are shown for endogenous-valid, endogenous-invalid, exogenous-valid, exogenous-invalid task variations.}
    \label{fig:posner_coa}
\end{figure}

To examine the effect that the CTOA interval plays in reaction time, the previous trial variations were performed for various CTOA lengths. The average reaction times are shown in Fig. \ref{fig:posner_coa}.

\subsection{Action Signals from Bottom-up Attention}

Since action can be determined from free-energy optimization, overt attention in the form of eye saccades or camera orientation changes can be as well implemented.
Here we examined focus reach times for two action-update contributions:
\begin{itemize}
    \item Top-down proprioceptive action signals: \(- \frac{\partial \boldsymbol{\tilde{s}}}{\partial a}^T{\boldsymbol{\tilde \Pi}_s} \boldsymbol{\tilde e}_s\) -- these are determined from the prediction error of the proprioceptive channel, between the proprioceptive input and current proprioceptive beliefs (which are attracted to higher intentions)
    \item Bottom-up visual precision action signals: \(\frac{1}{2} \mathrm{Tr}\left[\boldsymbol{\tilde{\Pi}_s^{-1}}\frac{\partial\boldsymbol{\tilde{\Pi}_s}}{\partial \boldsymbol{\tilde s}}\right]\frac{\partial \boldsymbol{\tilde{s}}}{\partial a} - \frac{1}{2} {\boldsymbol{\tilde e}_{s}^T\frac{\partial\boldsymbol{\tilde{\Pi}_s}}{\partial \boldsymbol{\tilde s}}\boldsymbol{\tilde e}_s}\frac{\partial \boldsymbol{\tilde{s}}}{\partial a}\) -- these are determined through the bottom-up derivative of the precision matrix. Since the action update is dependent only on the sensory input, only the second half of (\ref{eq:little_pi}) contributes to the action update.
\end{itemize}

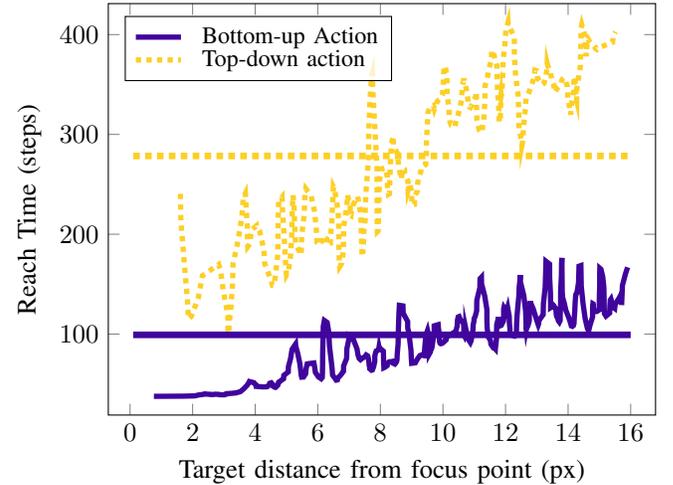
\begin{figure}[!t]
    \centering
    \ifthenelse{\usetikz=1}{
        \begin{tikzpicture}
            \begin{axis}[
                height=\figheight,
                width=\linewidth,
                xlabel={Target distance from focus point (px)},
                ylabel={Reach Time (steps)},
                grid=none,
                enlargelimits=0.05,
                legend style={
                    fill=white,  
                    fill opacity=\legendopacity,
                    text opacity=1,
                    font=\small,  
                    cells={anchor=west},  
                    row sep=-3pt,  
                    column sep=5pt  
                },
                legend pos=north west
            ]
                \addplot[dark_blue, mark=none, line width=2pt] table [col sep=comma, x=bottomx, y=bottomy] {data/posner_action.csv};
                \addlegendentry{Bottom-up Action};
                \addplot[dark_blue, line width=2.5pt, forget plot] coordinates {(0.1,99.21) (16,99.21)};
                
                \addplot[yellow,dotted, mark=none, line width=2pt] table [col sep=comma, x=topx, y=topy] {data/posner_action.csv};
                \addlegendentry{Top-down action}
                \addplot[yellow, dotted, line width=2.5pt, forget plot] coordinates {(0.1,278.42) (16,278.42)};
            \end{axis}
        \end{tikzpicture}
    }{
        \includegraphics[width=1.0\linewidth]{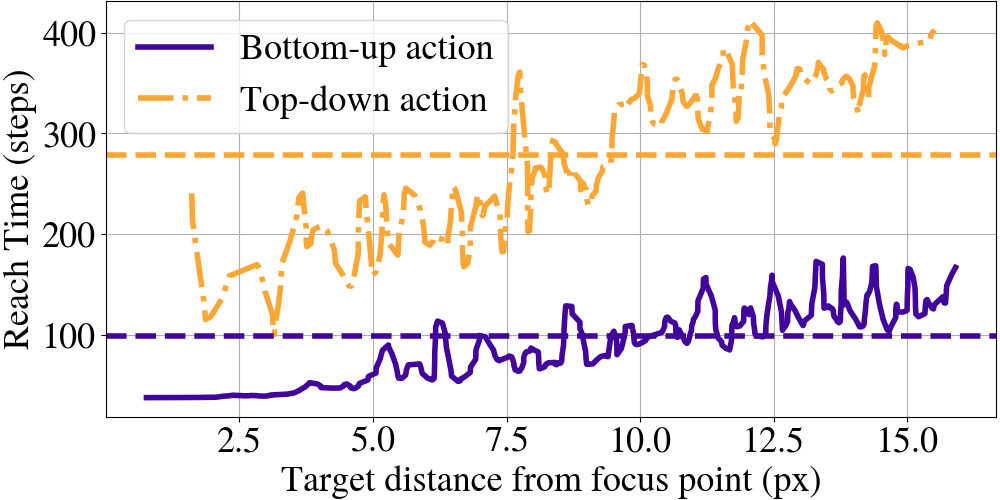}
    }
    \caption{Reach times and their averages for different initial target distances.}
    \label{fig:action}
\end{figure}

The trials start with a 10-step initialization interval, after which the target appears at a random position in the agent's field of view. The trial is finished when the agent successfully focuses the target at the center of its field of view. The reach times as a function of the initial target distance can be seen in Fig. \ref{fig:action}.

\section{Discussion} \label{discussion}

Our proposed model was tested on exogenous, endogenous, valid and invalid variations of the Posner paradigm, as well as on a simple target reach task. 
It captures the effects of both endogenous and exogenous attention, as well as the impact of cue validity, along with overt attention behaviors in involuntary actions, all of which have been observed in location-based models and human experimental data.
From the results in Fig. \ref{fig:posner_dist} we can conclude the following:
\begin{itemize}
    \item On average, valid cues produce faster reaction times than invalid ones\cite{posner1, posner2}. This can be explained by the location-based encoding of the target and the location-based covert focus in the visual image. This produces a spotlight effect suggested in location-based models of attention\cite{vecera,egly,reppa}. An invalid cue causes a greater shift of the "spotlight" upon target onset, thus increasing reaction time.
    \item Bottom-up exogenous cues produce faster reaction times than top-down endogenous cues \cite{jonides,cheal}. Bottom-up exogenous cues by error gradients through the VAE decoder are faster and require no interpretation in higher intentional areas, unlike top-down endogenous cues which require intentional interpretation of symbolic cues to update target belief.
    \item reaction times for every trial variation increase as target eccentricity increases \cite{klein, downing, carrasco}. This is a result of the location-based object encoding, as well as the shape of the RBF used for the precision matrix.
\end{itemize}

Regarding the shifts in covert attention demonstrated in Fig. \ref{fig:posner_shift}, covert attentional focus is much faster to update than the belief over the target's location, in the case of both endogenous and exogenous cues. 
This mirrors the findings that covert attentional shifts occur quickly, before conscious perception of target\cite{posner2,peterson} or active overt shifts in attention\cite{walker, kauffmann}.

Fig. \ref{fig:posner_coa} illustrates the effect of different CTOA intervals on reaction times, with invalid cues leading to faster reaction times than valid ones in the exogenous variation after longer CTOA intervals ($\sim$350 steps), and in the endogenous variation at a slightly later stage. 
Although not explicitly modeled, this behavior is similar to an attentional mechanism of inhibition of return (IOR) \cite{klein,reppa}, where a previously cued visual area becomes attentionally supressed after longer CTOA intervals in exogenous cues. 
Since this was not explicitly modeled, this model behavior will be examined in future work.

Overt visual attention in the form of camera orientation action signals was examined in a simple target reach task. 
The results in Fig. \ref{fig:action} show that bottom-up overt orienting is overall faster than top-down intentional orienting, which is explained by the sensitivity of the precision to red objects (or any predetermined visual object of interest, like faces \cite{kauffmann}). 
This is similarly reflected in how reaction time changes with distance.
Both forms of orienting exhibit an increasing trend in reaction time as distance increases; however, top-down orienting shows a steeper rise, indicating a greater sensitivity to distance compared to bottom-up orienting.
Although bottom-up overt orienting is faster, it can only effectively orient to one point in the visual area, while top-down overt orienting can handle multiple objects through multiple flexible intentions (at the cost of speed).
We leave multiple-object overt attention for future work.

\section{Conclusion}\label{conclusion}

In this paper, we have proposed an active inference model of covert and overt visual attention. 
The proposed model successfully demonstrates known attentional phenomena and mechanisms in the context of the Posner cueing task and a simple active orienting task. 
It shows that valid cues produce faster reaction times than invalid cues, and that exogenous cues produce faster reaction times than endogenous cues. 
The model also successfully demonstrates location-based attention, with reaction times increasing with target eccentricity. 
Although not modeled, the developed model exhibits behavior similar to inhibition of return, with previously cued areas becoming suppressed after a certain cue-target onset asynchrony interval.

Future work will investigate this emergence of inhibition of return, as well as extend the model with multiple possible targets/intentions to further test object-based and location-based effects. 
Overt saccades will also be examined further, with a focus on varying attraction to different objects.
We plan to further develop and test this framework as a model of perception, learning, and action in autonomous robots.

\balance

\bibliographystyle{ieeetr}
\bibliography{ref}

\appendix

\subsection{Implementation of gradients}
\label{appendix}

The gradients with respect to beliefs, action and sensory data given in (\ref{eq:belief_update}) and (\ref{eq:action_update}) depend on the different implementations of system dynamics, generative models, sensory precisions and the type of sensory data:

\begin{itemize}
    \item \(\frac{\partial \boldsymbol{\tilde f}}{\partial \boldsymbol{\tilde \mu}}\) : The gradient of the system dynamics function defined in (\ref{eq:system_dynamics}) w.r.t. the belief $\boldsymbol{\mu}$ is fairly simple, seeing as it is defined as an affine transformation of the belief.
    \item \(\frac{\partial \boldsymbol{\tilde g}}{\partial \boldsymbol{\tilde \mu}}\) : The gradients of the generative models w.r.t. the belief for the proprioceptive and interoceptive models are simple identity matrices. However, the gradient of the visual generative model is the gradient of the VAE decoder computed by backpropagation.
    \item \(\frac{\partial\boldsymbol{\tilde{\Pi}_s}}{\partial \boldsymbol{\tilde \mu}}\) : Since the sensory precision matrix is assumed to be diagonal, this greatly simplifies calculation of the gradients \(\frac{\partial \pi_i}{\partial \boldsymbol{\mu}}\) for each pixel \(i\) from the individual precision functions  \(\pi_i(\boldsymbol{\mu},\boldsymbol{s})\). The sensory precision gradient \(\frac{\partial\boldsymbol{\tilde{\Pi}_s}}{\partial \boldsymbol{\tilde \mu}}\) is a tensor of shape \(L\times L \times M\).
    \item \(\frac{\partial\boldsymbol{\tilde{\Pi}_\mu}}{\partial \boldsymbol{\tilde \mu}}\) : The optimization of system dynamics precisions \(\boldsymbol{\tilde{\Pi}_\mu}\) is left for future work, and they are assumed to be constant. Their gradients are therefore zero.
    \item \(\frac{\partial\boldsymbol{\tilde{\Pi}_s}}{\partial \boldsymbol{\tilde s}}\) : The gradient is calculated in a way similar to \(\frac{\partial\boldsymbol{\tilde{\Pi}_s}}{\partial \boldsymbol{\tilde \mu}}\), with the gradient being a tensor of shape \(L\times L \times L\).
    \item \(\frac{\partial \boldsymbol{\tilde{s}}}{\partial a}\) : The inverse mapping from sensory data to actions is generally considered a ``hard problem'' \cite{Friston2010_2}. However, it is fairly simple in our case: the centroid of the color red is converted into pitch and yaw angles (assuming we know the intrinsic parameters of the camera model).
\end{itemize}

\subsection{Variational Autoencoder} \label{vae}

The encoder consists of a convolutional layer $3 \times 3$ (in channels: 3, out channels: 32), followed by four residual downsampling blocks (32→64, 64→128, 128→256, 256→512). 
A fully connected layer maps the 512-dimensional feature vector to 64, followed by another producing a $2\times 8$-dimensional latent space output. 
The decoder mirrors this structure, with a fully connected layer expanding 8 to 64, reshaped into a $512 \times H/16 \times W/16$ feature map, followed by four residual upsampling blocks (512→256, 256→128, 128→64, 64→32) and a final $3 \times 3$ convolutional layer (32 output).
The VAE was implemented and trained in \textit{pytorch} on 240,000 $32\times32\times3$ images randomly generated in the Gazebo simulator.
The latent space was disentangled with manual encodings of sphere's image coordinates for each of the training images.

\subsection{Sensory Data and Belief Shape} \label{sensory_belief}
The three different kinds of sensory input are as follows:
\begin{itemize}
    \item Proprioceptive: pitch and yaw angles of the camera's orientation in the simulator, expressed in radians.
    \item Visual: a $32\times32\times3$ RGB image captured by the simulated camera model.
    \item Symbolic cue: a floating-point array with two elements, containing the image coordinates that cue where the target may appear.
\end{itemize}

The belief is a concatenation of the following elements:
\begin{itemize}
    \item Symbolic cue belief: two elements that mirror the sensory input for the symbolic cue
    \item Proprioceptive belief: two elements that mirror the sensory input for the pitch and yaw angles
    \item Visual encoding belief: the visual encoding used by the decoder to generate visual predictions. The first three elements encode the sphere's position and presence, while the rest are free latent variables
    \item Covert focus belief: represents the center and amplitude of the RBF used in the calculation of the sensory precision.
\end{itemize}

\end{document}